\documentclass[10pt,twocolumn,letterpaper]{article}

\usepackage{cvpr}
\usepackage{times}
\usepackage{epsfig}
\usepackage{graphicx}
\usepackage{amsmath, amssymb}
\usepackage{amssymb}
\usepackage{multirow}
\usepackage{subcaption}
\usepackage{multirow,bigdelim}
\usepackage{booktabs}

\usepackage[breaklinks=true,bookmarks=false]{hyperref}

\cvprfinalcopy 

\def\cvprPaperID{****} 

\begin{document}

\title{Dynamic Fusion with Intra- and Inter-modality Attention \\Flow for Visual Question Answering}

\author{Peng Gao\textsuperscript{1}, \ Zhengkai Jiang\textsuperscript{3}, \ Haoxuan You\textsuperscript{4},\\ Pan Lu\textsuperscript{4},\ Steven Hoi \textsuperscript{2},\ Xiaogang Wang \textsuperscript{1}, \ Hongsheng Li \textsuperscript{1} \\ \textsuperscript{1}CUHK-SenseTime Joint Lab, The Chinese University of Hong Kong\\
\textsuperscript{2}Singapore Management University \
\textsuperscript{3}NLPR, CASIA \
\textsuperscript{4}Tsinghua University\\
\textit {\{1155102382@link, xgwang@ee, hsli@ee\}.cuhk.edu.hk}\\
}

\author{Peng Gao\textsuperscript{1}, \ Zhengkai Jiang\textsuperscript{3}, \ Haoxuan You\textsuperscript{4},\\ Pan Lu\textsuperscript{4},\ Steven Hoi \textsuperscript{2},\ Xiaogang Wang \textsuperscript{1}, \ Hongsheng Li \textsuperscript{1} \\ \textsuperscript{1}CUHK-SenseTime Joint Lab, The Chinese University of Hong Kong\\
\textsuperscript{2}Singapore Management University \
\textsuperscript{3}NLPR, CASIA \
\textsuperscript{4}Tsinghua University\\
\textit {\{1155102382@link, xgwang@ee, hsli@ee\}.cuhk.edu.hk}\\
}

\maketitle

\begin{abstract}
Learning effective fusion of multi-modality features is at the heart of visual question answering. We propose a novel method of dynamically fusing multi-modal features with intra- and inter-modality information flow, which alternatively pass dynamic information between and across the visual and language modalities. It can robustly capture the high-level interactions
between language and vision domains, thus significantly improves the performance of visual
question answering. We also show that the proposed dynamic intra-modality attention flow conditioned on the other modality can
dynamically modulate the intra-modality attention of the target modality, which is vital for multimodality feature fusion. Experimental evaluations on the VQA 2.0 dataset show that the proposed method achieves state-of-the-art VQA performance. Extensive
ablation studies are carried out for the comprehensive analysis of the proposed method.
\end{abstract}

\section{Introduction}
Visual Question Answering~\cite{antol2015vqa} aims at automatically answering a natural language question related to the contents of a given image.
It has extensive applications in practice, such as assisting blind people~\cite{gurari2018vizwiz} and education of young children,
and therefore become a hot research topic recently.
The performance of Visual Question Answering (VQA) has been substantially improved in recent years thanks to three lines of works.
Firstly, better visual and language feature representations are at the core for boosting VQA performance.
The feature learning capability from VGG~\cite{simonyan2014very}, ResNet~\cite{he2016deep}, FishNet~\cite{sun2018fishnet} to the recent bottom-up \& top-down features~\cite{anderson2018bottom} increases
the VQA performance significantly.
Secondly, different variants of attention mechanisms~\cite{xu2015show} can adaptively select important features which can help deep learning achieve better
recognition accuracy. Thirdly, better multi-modality fusion approaches, such Bilinear Fusion ~\cite{gao2016compact}, MCB ~\cite{fukui2016multimodal} and
MUTAN ~\cite{Ben-younes_2017_ICCV}, have been proposed
for better capturing the high-level interactions between language and visual features.

Despite being studied extensively, most existing VQA approaches focus on learning inter-modality relations between visual and language features.
Bilinear feature fusion approaches ~\cite{gao2016compact} focus on capturing the higher order relations between language and visual modalities
by feature outer product. Co-attention~\cite{xiong2016dynamic, lu2016hierarchical, li2017identity} or bilinear attention-based approaches ~\cite{kim2018bilinear}
learn the inter-modality relations between
word-region pairs to identify key pairs for question answering. On the other hand, there exist computer vision and natural language
processing algorithms focusing on learning intra-modality relations. Hu \etal ~\cite{hu2018relation} proposed to explore intra-modality object-to-object
relations to boost object detection accuracy. Yao \etal \cite{yao2018exploring, liu2018show} modeled intra-modality object-to-object relations for improving image
captioning performance. In the recently proposed BERT algorithm \cite{devlin2018bert} for natural language processing, intra-modality word relations
are modelled by self-attention mechanism to learn state-of-the-art word embedding.
However, the inter- and intra-modality relations were never jointly investigated in a unified framework for solving the VQA problem.
We argue that, for the VQA problem, the intra-modality relations within each modality is
complementary to the inter-modality relations, which were mostly ignored by existing VQA methods.
For instance, for the image modality, each image region should obtain information not only from its associate words/phrases in the question
but also from related image regions to infer the answer of the question.
For the question modality, better understanding of question can be acquired by inferring other words.
Such cases motivate us to propose a unified framework for modelling
both inter- and intra-modality information flow.

\begin{figure*}[t]
        \begin{center}
                \includegraphics[width=\linewidth]{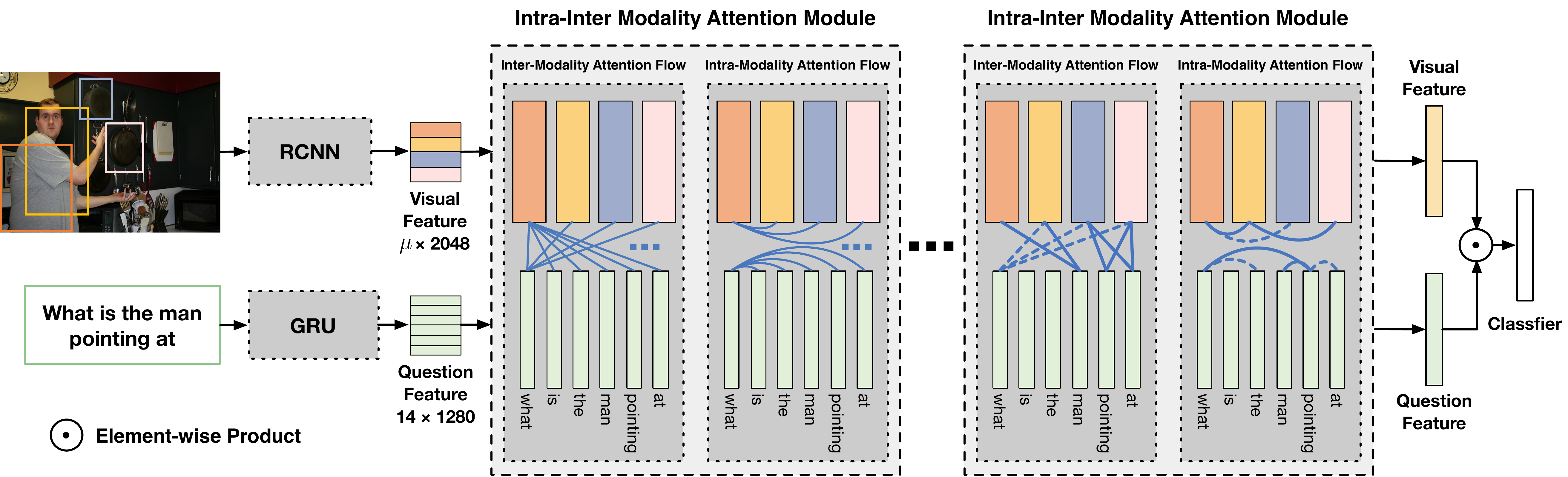}
        \end{center}
        \caption{Illustration of the proposed Dynamic Fusion with Intra- and Inter-modality Attention Flow (DFAF) for visual question answering. Each DFAF module contains one Inter-Modality Attention Flow and one of Intra-Modality Attention Flow Module. Stacking several blocks of DFAF can help the network gradually focus on important image regions , question words and the latent alignments.}
        \label{fig:overall}
\end{figure*}

To overcome the limitations, we propose a novel Dynamic Fusion with Intra- and Inter-modality Attention Flow (DFAF) framework for efficient
multi-modality feature fusion to accurately answer visual questions. The overall diagram is shown in Figure~\ref{fig:overall}. Our DFAF framework integrates cross-modal
self-attention and cross-modal co-attention mechanisms to achieve effective information flows within and between the image and
language modalities.
Given visual and question features encoded by deep neural networks, the DFAF framework first generates inter-modality
attention flow (InterMAF) to pass information between image and language.
In the InterMAF module, visual and language features generate
a joint-modality co-attention matrix.
Each visual region would select question features according to the joint-modality co-attention matrix and vice versa.
The InterMAF module fuses and updates each image region and each word's features according to the attention-weighted
information flows from the other modality. Following the InterMAF module, DFAF calculates the dynamic intra-modality attention
flow (DyIntraMAF) for passing information flows within each modality to capture the complex intra-modality relations.
Visual regions and sentence words generate self-attention weights and aggregate attention-weighted information
from other instances in the same modality. More importantly, although the information are only propagated within
the same modalities, information of the other modality is considered and used to modulate intra-modality attention weights and flows.
With such an operation, the attention flows within each modality are dynamically conditioned on the other modality
and is the key difference compared with existing intra-modality
message passing methods on object detection \cite{hu2018relation} and image captioning \cite{yao2018exploring}.
DyIntraMAF is shown to be substantially better than its
variant using only internal information for intra-modality
information flow and is the key to the success of the proposed framework.
We alternatively use  InterMA and DyIntraMA modules to create the basic blocks of
the DFAF. Multiple stacks of DFAF blocks are shown to further improve the VQA performance.

Our contributions can be summarized into threefold.
(1)
A novel Dynamic Fusion with Intra- and Inter-modality Attention Flow (DFAF) framework is proposed for multi-modality fusion by
interleaving intra- and inter-modality feature fusion. Such a framework for the first time integrates inter-modality
and dynamic intra-modality information flow in a unified framework for tackling the VQA task.
(2)
Dynamic Intra-modality Attention Flow (DyIntraMAF) module is proposed for generating effective attention
flows within each modality, which are dynamically conditioned on the information of the other modality. It is one of the core novelties of our proposed framework.
(3)
Extensive experiments and ablation studies are performed to examine the effectiveness of the proposed DFAF framework, in which state-of-the-art VQA performance is achieved by our proposed DFAF framework.

\section{Related Work}

\textbf{Representation learning for VQA.}
The recent boost of VQA performance is due to the success of deep representation learning.
In the early stage of VQA methods, the VGG~\cite{simonyan2014very} network was commonly used. With the introduction of  ResNet~\cite{he2016deep}, the VQA community shift to ResNet networks, which outperform VGG by large margins. Recently, the bottom-up and top-down network~\cite{anderson2018bottom}
derived from faster RCNN~\cite{ren2015faster} are shown to be suitable for VQA and image captioning tasks. Feature learning is an essential component for the development of VQA algorithms.

\textbf{Bilinear Fusion for VQA.}
Solving VQA requires understanding of visual and language contents as well as the relation between them. In early VQA methods, simple concatenation or
element-wise multiplication~\cite{zhou2015simple} between visual and language are used for cross-modal feature fusion.
To capture the high-level interactions between the two modalities, Bilinear Fusion~\cite{gao2016compact} has been proposed to adopt bilinear pooling to fuse features from the two modalities. To overcome the limitation of high computational cost of bilinear pooling, many approximated fusion methods, including
MCB~\cite{fukui2016multimodal}, MLB~\cite{kim2016hadamard} and MUTAN~\cite{Ben-younes_2017_ICCV}, were proposed, which have shown better performance than bilinear fusion~\cite{gao2016compact} with much fewer parameters.

\textbf{Self-attention-based methods.}
The attention mechanism in deep learning tries to mimic how human vision works. By automatically ignoring irrelevant information from the data, neural networks can selectively focus on important features. This approach has achieved great success in Natural Language Processing (NLP)~\cite{bahdanau2014neural}, image captioning~\cite{xu2015show} and VQA~\cite{zhu2017structured}. There are many variants of the attention mechanism.
Our approach are mainly motivated by self-attention and co-attention based methods. The self-attention mechanism~\cite{vaswani2017attention} transforms features into query, key and value features. The attention matrix between different features are then calculated by the inner product of query and key features. After acquiring the attention matrix, features are aggregated as the attention-weighted summation of the original features. Motivated by the self-attention mechanism, many vision tasks' performances were improved significantly. Non-local neural network~\cite{wang2017non} proposed a non-local  module for aggregating information between different frames within one video and achieved state-of-the-art performance in video classification. Relation Network learn~\cite{hu2018relation} the relationship between object proposals by adopting the self-attention mechanism. The in-place module can boost Faster RCNN ~\cite{ren2015faster} and Non-Maximum-Suppression (NMS) performance.

\textbf{Co-attention-based methods.}
The co-attention based~\cite{xiong2016dynamic, lu2016hierarchical} vision and language methods model the interactions across the two modalities.
For each word, every image region features are aggregated to the word according to the co-attention weights. The co-attention mechanism has been widely used in NLP and VQA tasks. In~\cite{nguyenimproved}, Dense Symmetric Co-attention (DCN) has been proposed. It achieved state-of-the-art performances on VQAv1 and VQAv2 datasets without using any bottom-up and top-down features. The success of DCN is due to dense concatenation~\cite{huang2017densely} of symmetric co-attention.

\textbf{Other works for language and vision tasks.}
Beyond above mentioned methods, many algorithms have also been proposed for fusion of cross-modal language and visual features.
Dynamic Parameter Prediction~\cite{noh2016image} and Question-guided Hybrid Convolution~\cite{gao2018question}
utilized dynamically predicted parameters for feature fusion. Adaptive attention~\cite{lu2017knowing} introduced a visual
sensual which can skip attention during image captioning. Structured attention~\cite{kim2017structured} adopted the MRF model over attention maps for better modelling better spatial attention distributions. Locally weighted deformable neighbours~\cite{Jiang_Gao_Guo_Zhang_Xiang_Pan_2019} proposed to predict offset and modulation weight.

\section{Dynamic Fusion with Intra- and Inter-modality Attention Flow for VQA}

\subsection{Overview}
The proposed approach consists of a series of DFAF modules. The whole pipeline is illustrated at figure~\ref{fig:overall} Visual and language features between the two modalities are first weighted with the co-attention mechanism and aggregated between the modalities to each image region and each word by the proposed Inter-modality Attention Flow (InterMAF) module, which learns the cross-modal interactions between the image regions and question words. Following the inter-modality module, to model the relationships within each modality, \ie, word-to-word relations and region-to-region relations, the Dynamic Intra-modality Attention Flow (DyIntraMAF) module is adopted. It weights words and regions within each modality and aggregates their features to the words and regions again, which could be viewed as passing information flows within each modality. Importantly, in our proposed intra-modality module, the attention flows are dynamically conditioned on the information from the other modality, which is a key difference compared with existing self-attention based methods. Such InterMAF and DyIntraMAF modules could be stacked multiple times to pass the information flows among words and regions iteratively to model the latent alignments for visual question answering.


\subsection{Base visual and language feature extraction}

To obtain base visual and language features, we extract image features from bottom-up \& top-down attention model~\cite{anderson2018bottom}.
The visual region features are obtained from a Faster RCNN~\cite{ren2015faster} model pretrained on Visual Genome~\cite{krishna2017visual} dataset. For each image, we extract 100 region proposals and its associated region features. Given the input image $I$, the obtained region visual features are denoted as $R \in \mathbb{R}^{ {\mu} \times {2048} } $, where the $i^\text{th}$ region feature is denoted as $r_i  \in \mathbb{R}^{2048}$ and there are $\mu$ object regions in total. The object visual features are fixed during training.

We adopt GLoVe word embeddings~\cite{pennington2014glove} as the inputs of the Gated Recurrent Unit (GRU)~\cite{chung2014empirical} for encoding question word features. Given the question $Q$, we obtain word-level features $E \in \mathbb{R}^{ {14} \times {1280}} $ from GRU, where the $j^\text{th}$ word feature is denoted as $e_j \in \mathbb{R}^{1280}$ and all questions are padded and truncated to the same length 14.

The obtained visual object region features $R$ and question features $E$ could be denoted as
\begin{align}
     R &= \textnormal{RCNN}(I; \theta_\text{RCNN}), \\
     E &= \textnormal{GRU} (Q; \theta_\text{GRU}).
\end{align}
where visual feature parameters $\theta_\text{RCNN}$ are fixed while question features $\theta_\text{GRU}$ are learned from scratch and updated together when training our proposed framework.

\subsection{Inter-modality Attention Flow}
The Inter-modality Attention Flow (InterMAF) module as shown in Figure~\ref{fig:overall} first learns to capture the importance between each pair of visual region and word features. It then passes information flows between the two modalities according to the learned importance weights and aggregate features to update each word feature and image region feature. Such an information flow process is able to identify cross-modal relations between visual regions and words.

Given visual region and word features, we first calculate the association weights between every pair of visual region and word.
Each visual region and word features are first transformed into query, key and value features following~\cite{shaw2018self, xu20192}, where
the transformed region features are denoted as $R_K$, $R_Q$, $R_V \in \mathbb{R}^{ {\mu} \times dim}$; Transformed word features are denoted as $E_K$, $E_Q$ and $E_V \in \mathbb{R}^{ {14} \times dim }$,
\begin{align}
	&R_K = \text{Linear}({R; \theta_{RK}}),  &E_K = \text{Linear}({R; \theta_{EK}}), \\
	&R_Q = \text{Linear}({R; \theta_{RQ}}),  &E_Q = \text{Linear}({E; \theta_{EQ}}),\\
	&R_V = \text{Linear}({R; \theta_{RV}}),  &E_V = \text{Linear}({E; \theta_{EV}}). \label{eq:value}
\end{align}
where ``Linear'' denotes a fully-connected layer with parameter $\theta$, and $dim$ represents the common dimension of transformed features from both modalities.

By calculating the inner product $R_{Q} E_K^T$ between every pair of visual region feature $R_Q$ and word key feature $E_K$, we obtain the raw attention weights for aggregating information from word features to each of the visual features, and vice versa. After normalizing the raw weights with the square root of the dimension number and a softmax non-linearity function, we obtain the two sets of attention weights, $\textnormal{InterMAF}_{R \leftarrow E} \in \mathbb{R}^{ {\mu} \times {14} }$ for weighting information flow transmitted from words to image regions, and $\textnormal{InterMAF}_{R \rightarrow E} \in \mathbb{R}^{ {14} \times {\mu} }$ for weighting information flow transmitted from image regions to sentence words,
\begin{align}
     \textnormal{InterMAF}_{R \leftarrow E} = \textnormal{softmax}(\frac{R_Q E_K^T}{\sqrt{dim}}), \\
     \textnormal{InterMAF}_{R \rightarrow E} = \textnormal{softmax}(\frac{E_Q R_K^T}{\sqrt{dim}}).
\end{align}
The inner product values are proportional to the dimension of hidden feature space, thus need to be normalized by the square root of hidden dimension.
The softmax non-linearity function is applied row-wisely.

The two bi-directional InterMAF matrices capture the importances between every image region and word pairs. Take the $\textnormal{InterMAF}_{R \leftarrow E} $ for example, each row stands for the attention weights between one visual region and all word embeddings. Information from all word embeddings to this one image region feature could be aggregated as the weighted summation of the word value features $E_V$. We denote the information flows to update visual region features and word features by the InterMAF module as $R_{update} \in \mathbb{R}^{\mu \times dim}$ and $E_{update} \in \mathbb{R}^{14 \times dim}$, respectively,
\begin{align}
       R_{update} = \textnormal{InterMAF}_{R \leftarrow E} \times E_V, \\
       E_{update} = \textnormal{InterMAF}_{R \rightarrow E} \times R_V.
\end{align}
where $E_V$ and $R_V$ are the un-weighted information flows(value features) to update visual region features and word features in Eq. \eqref{eq:value}, and the two InterMAF matrices are used to weight such information flows.

After acquiring the updated visual and word features, we concatenate them with original visual features $R$ and word features $E$. A fully connected layer is utilized to transform the concatenated features into output features, 
\begin{align}
       R = \textnormal{Linear}([R, \,R_{update}]^T; \theta_{RT}), \\
       E = \textnormal{Linear}([E, \,E_{update}]^T; \theta_{ET}).
\end{align}

The output features by the InterMAF module would then be fed into the following Dynamic Intra-modality Attention Flow module for learning intra-modality information flows to further update the visual region and word features for capturing region-to-region and word-to-word relations.

\subsection{Dynamic Intra-modality Attention Flow}

\begin{figure}[t]
        \begin{center}
                \includegraphics[width=\linewidth]{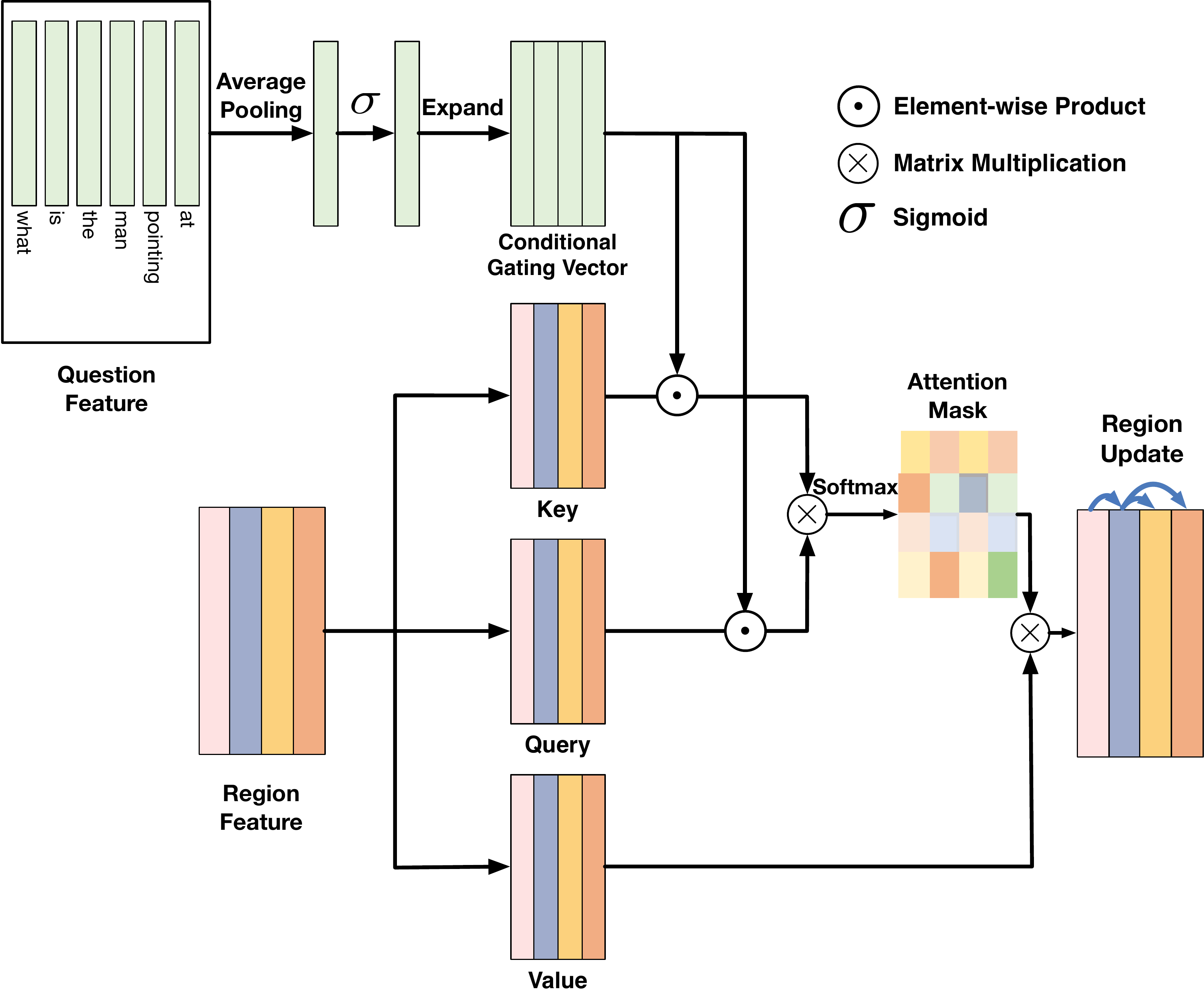}
        \end{center}
        \caption{Illustration of the proposed Dynamic Intra-Modality Attention Flow module. Only intra-modality attention flow within the visual modality conditioned on question are shown. By average pooling over question features, the conditional gating vector can be acquired for controlling the information flows among region features. Attention will focus on question related information flows. Row-wise softmax is applied to obtain the attention weight.}
        \label{fig:conditional}
\end{figure}

The input visual regions and word features of DyIntraMAF have encoded cross-modal relations between visual regions and words. However, we argue that relationships within each modality are complementary to the cross-modal relations and should be taken into account for improving the VQA accuracy. For example, for the question, ``who is above the skateboard ?'', the intra-modality module should relate the region above the skateboard and the skateboard region to infer the final answer. Therefore, we propose the Dynamic Intra-modality Attention Flow (DyIntraMAF) module for modelling such within-modality relations with a dynamic attention mechanism. The implementation of DyIntraMAF is illustrated at Figure~\ref{fig:conditional}.

The naive intra-modality matrices to capture the importance between regions and between words could be defined similarly to Eq. \eqref{eq:value} as,
\begin{align}
     \textnormal{IntraMAF}_{R \leftarrow R} = \textnormal{softmax}(\frac{R_Q R_K^T}{\sqrt{dim}}), \\
     \textnormal{IntraMAF}_{E \leftarrow E} = \textnormal{softmax}(\frac{E_Q E_K^T}{\sqrt{dim}}).
\end{align}
The dot products are utilized to estimate their within-modality importance between the same modality's query and key features.
Such weight matrices could then be used to weight the information flows transmitted within each modality.
Modelling within-modality relationships have been shown to be effective in object detection~\cite{hu2018relation}, image captioning
and BERT word embedding pretraining~\cite{devlin2018bert}.

However, the naive IntraMAF module only utilizes within-modality information for estimating the region-to-region and word-to-word importance. Some relations are important but could only be identified conditioned on information from the other modality. For instance, even for the same input image, relations between different visual region pairs should be weighted differently according to different questions. We therefore propose a Dynamic Intra-modality Attention Flow (DyIntraMAF) module to estimate within-modality relation importance conditioned on the information from the other modality.

To summarize the conditioning information from the other modality, we average pool the visual region features along the object-index dimension and the word features along the word-index dimension. The average pooled features of both modalities are then transformed to a $dim$-dimensional feature vector to match the dimension of the query and key features $R_Q, R_K, E_Q, E_K$. The $dim$-dimensional feature vector of each modality is then processed by a sigmoid non-linearity function $\sigma(\cdot)$ to generate channel-wise conditioning gates for the other modality,
\begin{align}
	G_{R \rightarrow E} = \sigma(\text{Linear}(\text{Avg\_Pool}(R); \theta_{RP})),\\
	G_{R \leftarrow E} = \sigma(\text{Linear}(\text{Avg\_Pool}(E); \theta_{EP})).
\end{align}
The query and key features from both modalities are then modulated by the conditional gates from the other modality
\begin{align}
	\hat{R}_Q = (1 +  G_{R \leftarrow E}) \odot R_Q, \,\,\, \hat{R}_K = (1 +  G_{R \leftarrow E}) \odot R_K, \nonumber\\
	\hat{E}_Q = (1 + G_{R \leftarrow E}) \odot  R_Q, \,\,\, \hat{E}_K = (1 +  G_{R \leftarrow E}) \odot E_K.
\end{align}
where $\odot$ denotes element-wise multiplication. Channels of query and key features would be activated or deactivated by channel-wise gates conditioned on the other modality. Such a design of the two gating vectors share the similar spirit with Squeeze and Excitation Network \cite{hu2017squeeze} and the Gated Convolution \cite{gehring2017convolutional}. The key difference is that the channel-wise gating vector is created based on cross-modal information.

The dynamic intra-modality attention flow matrices $\text{DyIntraMAF}_{R \leftarrow R} \in \mathbb{R}^{\mu \times \mu}$ and  $\text{DyIntraMAF}_{E \leftarrow E} \in \mathbb{R}^{14 \times 14}$ are then obtained by the gated query and key features to weight different within-modality relations,
\begin{align}
	\textnormal{DyIntraMAF}_{R \leftarrow R} = \textnormal{softmax}(\frac{\hat{R}_Q \hat{R}_K^T}{\sqrt{dim}}), \\
     \textnormal{DyIntraMAF}_{E \leftarrow E} = \textnormal{softmax}(\frac{\hat{E}_Q \hat{E}_K^T}{\sqrt{dim}}).
\end{align}

Visual region and word features are then updated by the weighted value features $R_V$ and $E_V$ via residual,
\begin{align}
       R = \textnormal{Linear}(R +  R_{update}; \theta_{RD}), \\
       E = \textnormal{Linear}(E +  E_{update}; \theta_{ED}).
\end{align}
where
\begin{align}
       R_{update} = \textnormal{DyIntraMAF}_{R \leftarrow R} \times R_V, \\
       E_{update} = \textnormal{DyIntraMAF}_{E \leftarrow E} \times E_V.
\end{align}

Note that here we only make key and query features conditioned on the other modality to adaptively weight within-modality information flows. In our ablation studies, we observe that the proposed DyIntraMAF module by a large margin outperforms the naive IntraMAF module.

\subsection{The Framework with Intra- and Inter-modality Attention Flow}

In this section, we introduce how to integrate intra- and inter-modality attention flow modules into our proposed framework. The whole pipeline is illustrated in Figure~\ref{fig:overall}.The proposed framework first extracts visual region features and word features from the input image and question by utilizing the Faster RCNN and GRU models, respectively. Faster R-CNN model weights are fixed during training our proposed framework, while GRU weights are updated with our framework from scratch.

After visual region features and word features being transformed into vectors of the same dimension by fully connected layers. The InterMAF module passes information flows between each pair of visual region and question word and aggregates updated features to each region and each word. Such aggregated features integrate information from the other modality to update the visual and word features according to the cross-modal relations.

Given the InterMAF outputs, the DyIntraMAF module is utilized for dynamically passing information flows within each modality. The visual region and word features would be updated again with information within the same modality via residual connections.

We use one InterMAF module followed by one DyIntraMAF module to form a basic block in our proposed DFAF framework. Multiple blocks could be stacked thanks to the feature concatenation and residual connection in the feature updating procedures. Very deep intra- and inter-modality information flows can be effectively trained with stochastic gradient descent. In addition, we utilize multi-head attention in practice.
The original features are split along channel dimensions into groups and different groups would generate parallel attentions to update visual and word features in different groups independently.

\subsection{Answer Prediction Layer and Loss Function}
After several blocks of feature updating by InterMAF and DyIntraMAF modules, we obtain the final visual region and word features encoding inter-modality and intra-modality relations for VQA.
By average pooling over region features and over word features, we could obtain discriminative representations for image and question, respectively. Such features could then be fused via either feature concatenation, or feature element-wise product, or feature addition to obtain fused features. We experiment with the three fusion approaches in which the element-wise product between visual and language representations achieves the best performance with a trivial margin.

Similar to state-of-the-art VQA approaches, we treat VQA as a classification problem. The fused multi-modal features are transformed into
a probability vector by a 2-layer multi-layer perceptron with ReLU non-linearity function between the layers and a final softmax function. The ground-truth answers are extracted from annotated answers that appear for more than 5 times. Cross-entropy loss function is adopted as the objective function.
\section{Experiments}
\subsection{Datasets}
We used VQA version 2.0~\cite{balanced_vqa_v2} for our experiment. VQA dataset contains human annotated
question-answer pairs for images from Microsoft COCO dataset ~\cite{lin2014microsoft}.
VQA 2.0 is an updated of previous VQA 1.0 with much more annotations and less dataset bias.
VQA 2.0 is split into train, validation and test-standard sets.
Among test-standard test, 25\% are served as test-dev set. All questions types are divided into Yes/No, Number and other categories.
Train, validation and test-standard contains 82,783, 40,504 and 81,434 images, with 443,757, 214,354, 447,793 questions,respectively. Each question
contains 10 answers from different annotators. Answers with the highest frequency are treated as the ground-truth. Following previous
approaches, we perform ablation studies over the validation set and utilize the  train and validation splits for test.

\subsection{Experimental Setup}
Visual features of dimension 2048 are extracted from Faster R-CNN~\cite{ren2015faster} while word features are encoded into features of dimension 1280 by GRU\cite{chung2014empirical}. The visual features and word features are then embedded into 512 dimensions by a fully-connected layer, respectively. Inside InterMAF, features are transformed into 8 multi-head attention with 64 dimensions for each head. For
DyIntraMAF, the average pooled features from both modality are transformed into 512 dimensions by MLP followed by element-wise sigmoid
activation to obtain the conditioning gating vectors. They are then multiplied with 512 dimension
visual key and query features at every position of visual and word features for dynamic attention flows. Previous approaches achieve
significantly better results with sentinel and relative position information. However, sentinel and relative position do not affect
the performance of our method.

All fully connected layers have the same dropout rate 0.1. All gradients are clipped to 0.25. Batch size is set as 512.
Adamax optimizer ~\cite{kingma2014adam},
a variant of Adam, is used. The learning rate is set as $10^{-3}$ for the first 2 epoch, $2 \times 10^{-3}$ for the next 8 epochs and decayed by 1/4 for the rest
epochs.
Our method is implemented with Pytorch ~\cite{paszke2017automatic}.
All initilizations are Pytorch default initilization.

All ablation studies are conducted on the validation dataset, while train, validation datasets and extra visual genome dataset are combined for testing on test-dev.

\begin{table}[tb]
\small
\centering
\begin{tabular}{l  c  c}
\toprule
Component & Setting & Accuracy \\
\toprule
Bottom-up~\cite{anderson2018bottom} & Bottom-up & 63.37 \\
\hline
\multirow{3}{2cm}{Bilinear Attention~\cite{kim2018bilinear}} & BAN-1 & 65.36 \\
& BAN-4 & 65.81 \\
& BAN-12 & 66.04 \\
\hline
Default &DFAF-1 & 66.21 \\
\hline
\multirow{3}{*}{\# of stacked blocks} & DFAF-2 & 66.43 \\
 & DFAF-5 & 66.58 \\
 & DFAF-8 & \textbf{66.66} \\
\hline
\multirow{2}{*}{Attention type} & InterMAF only &  64.37 \\
 & IntraMAF only & 62.34 \\
 & DyIntraMAF only & 65.51 \\
 & \multirow{2}{2cm}{InterMAF + DyIntraMAF} & \textbf{66.21} \\
 &     &  \\
\hline
\multirow{2}{3cm}{Attention Direction inside InterMAF} & Parallel & 65.99  \\
 & $R \rightarrow E$, $E \rightarrow R$ &  \textbf{66.21} \\
 & $E \rightarrow R$, $R \rightarrow E$ &  66.19 \\
\hline
\multirow{2}{2cm}{Embedding dimension} & 512 & \textbf{66.21} \\
&  1024 & 65.89 \\
\hline
\multirow{3}{2cm}{Cross-model feature fusion} & Multiplication & \textbf{66.21} \\
& Addition & 66.11\\
& Concatenation & 66.14\\
\hline
\multirow{3}{*}{Visual Sentinel} & None & \textbf{66.21} \\
& 1 & 66.01\\
& 3 & 66.02\\
\hline
\multirow{2}{3cm}{Bounding Box Embedding} & None & \textbf{66.21} \\
& Absolute Position & 65.88\\
& Relative Position & 65.23\\
\hline
\multirow{1}{*}{Parallel Heads} & 1 head each 512 & 65.84\\
& 4 heads each 128 & 66.17\\
& 8 heads each 64 & \textbf{66.21} \\
\bottomrule
\end{tabular}
\caption{Ablation studies of our proposed DFAF on VQA 2.0 validation dataset. R stands for region features while E stands for word
         embedding features}
\label{tab:ablation}
\end{table}

\subsection{Ablation study of DFAF}
We perform extensive ablation studies on the VQA 2.0 validation dataset~\cite{balanced_vqa_v2}. The results are shown in Table~\ref{tab:ablation}. Our default setting only has 1 block of DFAF module. Region features with 2,048 dimensions are extracted
from the input image by Faster RCNN~\cite{ren2015faster}, word features with 1,024 dimensions are extracted by GRU~\cite{chung2014empirical}. By default, all modules inside DFAF has 512
dimensions. In the final fusion layer, feature multiplication is employed, which shows a trivial improvement. Visual sentinel~\cite{lu2017knowing}
and bounding box position embedding are also tested which give a slight drop in the final performance. 8 parallel attention heads with
dimensions 64 for each head is utilized in the default setting. 

We first investigate the influence of the number of stacked DFAF blocks. The default setting has one stack. As one can see from Table
~\ref{tab:ablation}, more stacks can improve the performance thanks to the residual connection~\cite{he2016deep}. Different from ResNet,
we do not employ any normalization~\cite{ioffe2015batch} technique during residual connection. The performance of single layer DFAF is comparable with BAN-12~\cite{kim2018bilinear}.

Then, we investigate the influence of attention type. The attention mechanism in Bottom up~\cite{anderson2018bottom} utilizes simple attention methods. Bilinear
attention network~\cite{kim2018bilinear} proposed a bilinear attention which learns the joint attention distribution between each word and region pairs. By adding
the InterMAF, performance can improve by 1\% because of the modelling the inter-modality relations between image regions and question words.
Integrating only the IntraMAF module would harm the performance because too many unrelated information flows hinder the learning process.
By adding dynamically conditioned information flow DyIntra MAF module, we achieve a 2.15\% performance improvement. By combining Intra- and Inter-modality attention flows, we significantly outperform the baseline~\cite{anderson2018bottom} by 2.83\% and previous state-of-the-art BAN-1~\cite{kim2018bilinear} by 0.85\%.

There are several orders for passing information within the InterMAF module, namely, parallel and sequential~\cite{xiong2016dynamic, lu2016hierarchical}. For parallel InterMAF, both region and word
features are updated at the same time. For the sequential information flow, we experiment with passing attention flow from regions to
words first, which updates word features, and then passing message from words to regions, which then update region features, and vice versa. We denote the first
sequential order as $R \rightarrow E$, $E \rightarrow R$, and the second one as $E \rightarrow R$, $R \rightarrow E$. Sequential update outperforms parallel update way, while the specific order does not matter.

Next, we perform ablation study on the influence of embedding dimension and cross-model feature fusion. 512 dimensions result in better performance than
1024 dimensions. For the fusion method, multiplication shows a slight better performance than feature addition and concatenation.

Visual sentinel~\cite{lu2017knowing, xiong2016dynamic} were utilized in many previous VQA methods, which was shown to increase the VQA accuracy. We treat sentinel as a general 512
dimension features and concatenate sentinel with all region and word features. Previous $\mu$ region features and 14 word features become
into $\mu + 1$ and 15 respectively. In our experiments, adding visual sentinel do not show improvement.

\begin{figure*}[t!]
 \footnotesize
 \begin{tabular}{c@{\hspace{-2mm}}c}
  &
  \begin{tabular}{c@{\hspace{2mm}}p{1.6in}@{\hspace{2mm}}p{1.6in}@{\hspace{2mm}}p{1.6in}}

   \textbf{IntraMAF weights}&
   \textbf{~~~~~~~~~~DyIntraMAF weights}&
   \textbf{~~~~~~~~~~DyIntraMAF weights}&
   \textbf{~~~~~~~~~~DyIntraMAF weights}\\

   \vspace{0.01in}
   \includegraphics[width=1.6in]{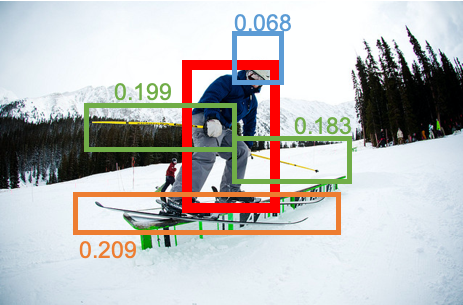}&
   \includegraphics[width=1.6in]{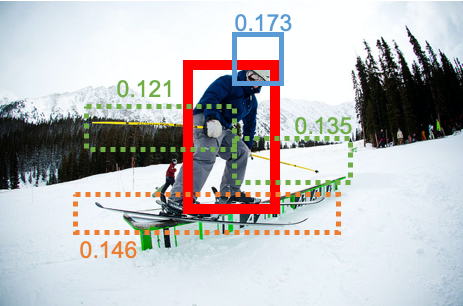}&
   \includegraphics[width=1.6in]{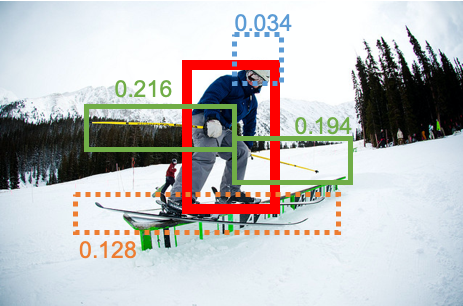}&
   \includegraphics[width=1.6in]{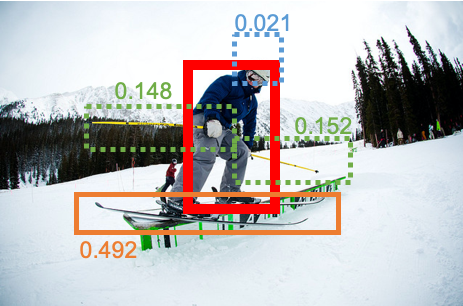}\\

   \textbf{Q}: {Is the \underline{\emph{skier}} wearing \underline{\emph{goggles}}?} \textbf{A}:{\color{red}No} &
   \textbf{Q}: {Is the \underline{\emph{skier}} wearing \underline{\emph{goggles}}?} &
   \textbf{Q}: {Is this \underline{\emph{skier}} using \underline{\emph{poles}}?} &
   \textbf{Q}: {What is the \underline{\emph{person}} standing on?}\\
   \vspace{0.01in}

   \textbf{Q}: {What is the \underline{\emph{person}} standing? } \textbf{A}:{\color{green}Skis}&
   \textbf{A}: {\color{green}Yes} &
   \textbf{A}: {\color{green}Yes} &
   \textbf{A}: {\color{green}Skis}\\

   \textbf{Q}: {Is this \underline{\emph{skier}}  using \underline{\emph{poles}}?} \textbf{A}:{\color{red}No}&
   &
   &
   \\

   \includegraphics[width=1.6in]{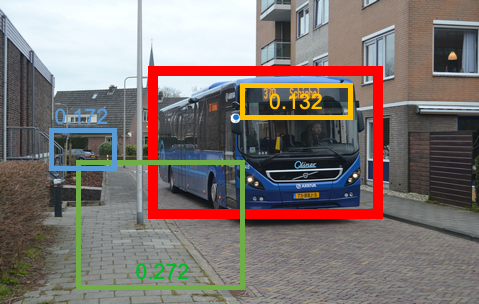}&
   \includegraphics[width=1.6in]{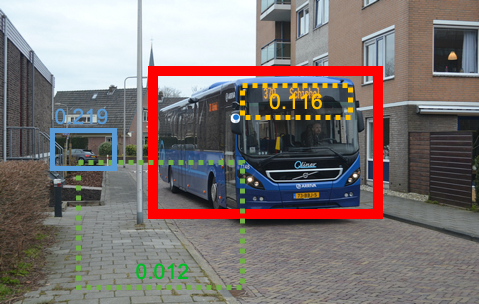}&
   \includegraphics[width=1.6in]{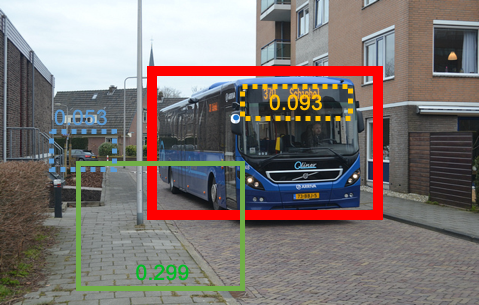}&
   \includegraphics[width=1.6in]{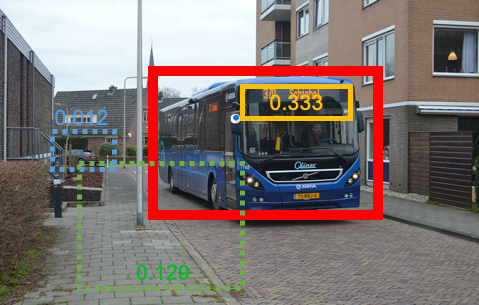}\\

   \textbf{Q}: {How many \underline{\emph{vehicles}} are in the street?} \textbf{A}:{\color{red}1} &
   \textbf{Q}: {How many \underline{\emph{vehicles}} are in street?} &
   \textbf{Q}: {Are there any \underline{\emph{people}} on street?} &
   \textbf{Q}: {Where will the bus \underline{\emph{go}}?}\\
   \vspace{0.01in}

   \textbf{Q}: {Are there any \underline{\emph{people}} on the street? } \textbf{A}:{\color{green}No}&
   \textbf{A}: {\color{green}2} &
   \textbf{A}: {\color{green}No} &
   \textbf{A}: {\color{red}School}\\

   \textbf{Q}: {Where will the bus \underline{\emph{go}}?} \textbf{A}:{\color{red}School}&
   &
   &
   \\
  \end{tabular}
 \end{tabular}
 \caption{Visualisation of IntraMAF and DyIntraMAF attention weights between central region(red) and other related regions. (Left) The IntraMAF module treats different questions equally and generate uninformative weight for different questions. (Right) The proposed DyIntraMAF module dynamically changes attention weights according to input questions.}
 \label{fig:vis}
\end{figure*}

The positions of bounding boxes were widely utilized as a part of image region features in previous methods. Absolute position embedding has been employed in
Transformer~\cite{vaswani2017attention}, BERT~\cite{devlin2018bert} and Gated CNN~\cite{gehring2017convolutional} in NLP. Relative position was adopted in relation network~\cite{hu2018relation} for object
detection. In our experiment, adding absolute or relative positions would drop the performance.

At last, we experiment on the influence of multi-head attention~\cite{vaswani2017attention}.
We keep the overall dimensions to be 512. 1, 4 and 8 attention heads are experimented. As can be seen in Table~\ref{tab:ablation},
8 attention can achieve better performance at the same number of parameters.

\subsection{Visualisation of the proposed Attention Flow Weights}
In Figure~\ref{fig:vis}, we visualise the intra-modality attention flow weights to analyse VQA model. The attention weights modulate information flow from contextual regions(orange, blue and green) to center region(red). The left column stands for the attention flow weights in the IntraMAF module. While the rest columns represent dynamic attention flow weights in the DyIntraMAF module. In the DyIntraMAF module, unrelated information flow are filtered out by question features and thus generate the correct answer.

\subsection{Comparison with State-of-the-arts methods}
Table~\ref{tab:vqa} shows the performance of our proposed algorithm trained with extra visual genome dataset and state-of-the-art methods on VQA.
Bottom up in Table ~\ref{tab:vqa} is the winner of VQA challenge 2017. This approach proposed to use features based on Faster RCNN~\cite{ren2015faster} instead of
ResNet~\cite{he2016deep}. Multi-modal Factorized High-order Pooling (MFH)~\cite{yu2018beyond} is a state-of-the-art bilinear pooling methods. Dense Co-Attention Network (DCN)~\cite{nguyenimproved} utilized dense stack of multiple
layers of Co-attention mechanism which significantly outperform previous methods with ResNet features. Counting methods~\cite{zhang2018learning} are good at counting
questions by utilizing the information of bounding boxes. Bilinear Attention Network (BAN)~\cite{kim2018bilinear} is a state-of-the-art approach on VQA 2.0 which has 12 stacked blocks of BAN modules. By utilising contextualised word embedding BERT~\cite{devlin2018bert}, the performance of VQA can be further boosted.

\begin{table}[tb]
\footnotesize
\centering
\begin{tabular}{l c c c c c}
\toprule
\multirow{2}{*}{Model} &      \multicolumn{4}{c}{test-dev} & \multicolumn{1}{c}{test-std} \\
\cmidrule(lr){2-5} \cmidrule(lr){6-6}
 & Y/N  & No. & Other & All & All\\
\hline
Bottom-up~\cite{anderson2018bottom} & 81.82 & 44.21 & 56.05 & 65.32  & 65.67 \\
MFH~\cite{balanced_vqa_v2} & n/a & n/a & n/a & 66.12 & n/a \\
DCN~\cite{nguyenimproved} & 83.51  & 46.61  & 57.26  & 66.87 & 66.97 \\
Counter~\cite{zhang2018learning} & 83.14 & 51.62 &  58.97 & 68.09 &  68.41 \\
MFH+Bottom-Up~\cite{balanced_vqa_v2} &  84.27 &  49.56 & 59.89 & 68.76 & n/a \\
BAN+Glove~\cite{kim2018bilinear} & 85.46 & 50.66 & \textbf{60.50} & 69.66 & n/a \\
\hline
DFAF(ours) &  \textbf{86.09}  & \textbf{53.32}& 60.49 &\textbf{70.22} &\textbf{70.34} \\
DFAF-BERT(ours) &  86.73  & 52.92& 61.04 &70.59 &70.81 \\
\bottomrule
\end{tabular}
\vspace{-6pt}
\caption{Comparison with previous state-of-the-art methods on VQA 2.0 test dataset.}
\label{tab:vqa}
\vspace{-9pt}
\end{table}

\section{Conclusions}
In this paper, we proposed a novel framework Dynamic Fusion with Intra- and Inter-modality Attention Flow (DFAF) for visual question answering. The DFAF framework alternatively passes information within and across different modalities based on an inter-modality and intra-modality attention mechanisms. The information flow inside visual features are
dynamically conditioned on the question features. Stacking multiple blocks of DFAF are shown to improve the performance of VQA.

\section*{Acknowledgment}
This work is supported in part by SenseTime Group Limited, in part by the General Research Fund through the Research Grants Council of Hong Kong under Grants CUHK14202217, CUHK14203118, CUHK14205615, CUHK14207814, CUHK14213616, CUHK14208417, CUHK14239816, and in part by CUHK Direct Grant.


{\small
\bibliographystyle{ieee_fullname}
\bibliography{egbib}
}

\end{document}